\documentclass[10pt,twocolumn,letterpaper]{article}

\usepackage{cvpr}
\usepackage{times}
\usepackage{epsfig}
\usepackage{graphicx}
\usepackage{amsmath}
\usepackage{amssymb}

\usepackage{makecell}
\usepackage{multirow}

\usepackage[latin9]{inputenc}
\usepackage[english]{babel}
\usepackage{caption}% <-- added
\usepackage{tabulary}
\usepackage[para]{threeparttable}
\usepackage{array,booktabs,longtable,tabularx}
\usepackage{ltablex}% <-- added
\usepackage{siunitx}% <-- added
\usepackage{caption}% <-- added
\usepackage[flushleft]{threeparttablex}
\usepackage{footnote}
\usepackage{tablefootnote}
\usepackage[hyphens,spaces,obeyspaces]{url}

\graphicspath{ {./images/} }
\makesavenoteenv{tabular}
\makesavenoteenv{table}

\usepackage[pagebackref=true,breaklinks=true,letterpaper=true,colorlinks,bookmarks=false]{hyperref}

\cvprfinalcopy % *** Uncomment this line for the final submission

\begin{document}

%%%%%%%%% TITLE
\title{Lightweight Network Architecture for Real-Time Action Recognition}

\author{Alexander Kozlov\\
Intel\\
{\tt\small alexander.kozlov@intel.com}
\and
Vadim Andronov\\
Intel\\
{\tt\small vadim.andronov@intel.com}
\and
Yana Gritsenko\\
Intel\\
{\tt\small yana.gritsenko@intel.com}
}

\maketitle
%\thispagestyle{empty}

%%%%%%%%% ABSTRACT
\begin{abstract}
In this work we present a new efficient approach to Human Action Recognition called Video Transformer Network (VTN). It leverages the latest advances in Computer Vision and Natural Language Processing and applies them to video understanding. The proposed method allows us to create lightweight CNN models that achieve high accuracy and real-time speed using just an RGB mono camera and general purpose CPU. Furthermore, we explain how to improve accuracy by distilling from multiple models with different modalities into a single model. We conduct a comparison with state-of-the-art methods and show that our approach performs on par with most of them on famous Action Recognition datasets. We benchmark the inference time of the models using the modern inference framework and argue that our approach compares favorably with other methods in terms of speed/accuracy trade-off, running at 56 FPS on CPU. The models and the training code are available\footnote{\href{https://github.com/opencv/openvino_training_extensions/tree/develop/pytorch_toolkit/action_recognition}{https://github.com/opencv/openvino\_training\_extensions/tree/develop/}  \href{https://github.com/opencv/openvino\_training\_extensions/tree/develop/pytorch\_toolkit/action_recognition}{pytorch\_toolkit/action\_recognition}}.

\end{abstract}

%%%%%%%%% BODY TEXT
\section{Introduction}

\begin{figure}[t!]
\includegraphics[width=8cm]{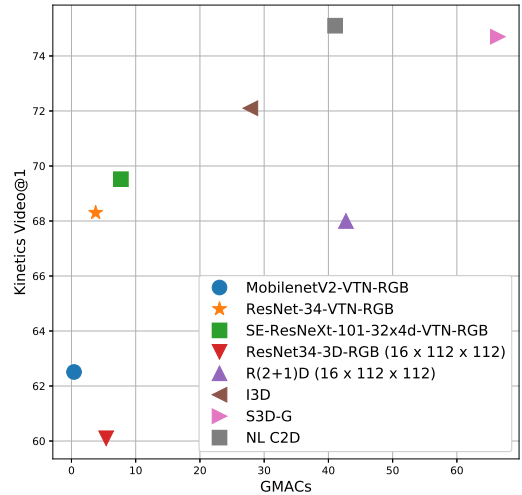}
\caption{\textbf{Accuracy vs complexity trade-off} for different methods on Kinetics-400 validation set. First three models are the variants of the proposed VTN method. ResNet-34 3D with a similar number of MAC (accepts smaller resolution inputs) is presented for comparison. We also included several state-of-the-art methods: I3D \cite{QuoVadis}, R(2+1)D \cite{CloserLook}, S3D-G \cite{xie2018rethinking}, NL-C2D \cite{wang2017non}.} 
\label{tradeoff}
\centering
\end{figure}

The latest advances in the Computer Vision domain are definitely related to the development of Deep Learning (DL) methods \cite{VGG, InceptionV1, ResNet} which show great results on many tasks such as Image Classification \cite{Imagenet} and Segmentation \cite{Cityscapes}, Object Detection \cite{VOC, MSCOCO}, etc. There is a tendency nowadays to create more and more sophisticated pipelines \cite{MaskR-CNN, PSPNet, OpenPose}, combining quite complex components which solve the task well but require a massive amount of calculations and power at the same time. On the other hand, since the times of AlexNet \cite{AlexNet} and VGG \cite{VGG} where a vanilla convolution was used as a basic building block, new lightweight primitives have been proposed~\cite{SqueezeNet, Xception, MobileNet, ShuffleNet}, allowing to reduce the theoretical complexity but retain or even improve the final accuracy. However, video-level tasks, such as Human Action Recognition, which is being discussed in this work, require to consider temporal structure of input data by aggregating information from multiple frames in order to solve action ambiguities (opening/closing the door). This inevitably incurs extra computational costs during inference of the model. Nevertheless, few studies \cite{massivelyParallel} pay attention to the complexity of the algorithm while maximizing accuracy. Therefore, creating a solution that can achieve high accuracy providing a fast inference speed would be a relevant task, especially in the case of low-power devices used for edge computing (at the edge).

Following this idea, we propose a lightweight architecture for AR which can run in real-time on a regular CPU, performing on par with heavy methods, such as 3D CNN \cite{C3D, QuoVadis, CloserLook}. In support of this, we provide a comparison (see Fig.~\ref{tradeoff} and Section~\ref{sota_compare}) of our model with the state-of-the-art methods and verify its accuracy on modern benchmarks, such as Kinetics~\cite{Kinetics}, UCF-101~\cite{UCF101}, and HMDB-51~\cite{HMDB}.
%Moreover, we tested our model on a reliable Jester dataset \cite{Jester} and got the first score, as well as verified it on more famous benchmarkes, like Kinetics \cite{Kinnetics} and UCF-101 \cite{UCF101}, in order to do a more fair comparison.

Shortly, our contributions can be summarized  as follows:
\begin{itemize}
    \item A new lightweight CNN architecture for real-time Action Recognition that achieves results comparable to state-of-the-art methods.
    \item Comparison of modern approaches to Action Recognition. 
    \item A method for improving the accuracy of an existing model by accommodating information from additional modality without a discernible increase in complexity.
    %\item The first score on the Jester \cite{Jester} dataset.
\end{itemize}
%-------------------------------------------------------------------------

\begin{figure*}[ht]
\includegraphics{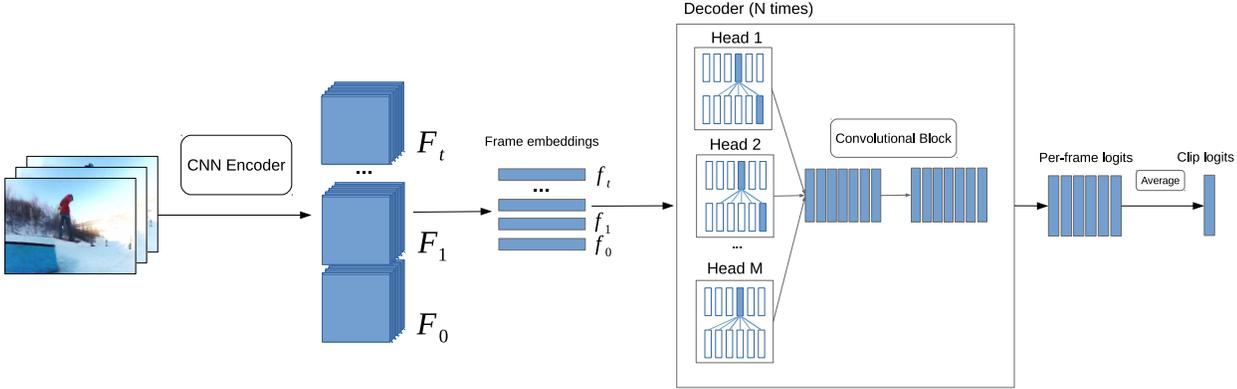}
\caption{Overview of the VTN architecture. $t$ input frames are fed to CNN encoder and global pooled to get frame embeddings. Then the decoder block (see in details in Fig.~\ref{attn_block}) is applied $N$ times. In the end, the clip logits are produced by averaging all frame logits.}
\label{fig:method_overview}
\end{figure*}

\section{Related Work} \label{RelatedWorks}
Currently, there are multiple methods that solve the AR problem with certain quality. %However, most of them are based on three general ideas. may be classified into three general categories. 

One of the examples is the two-stream framework that fuses information from spatial and temporal nets \cite{simonyan2014two_stream, Zisserman_Two_Stream_2016}. Spatial net uses RGB frame as input and represents an ordinary classification CNN working on a frame level, whereas temporal net receives multiple stacked optical flow (OF) frames. Calculating OF with traditional algorithms, such as TVL1~\cite{TVL1}, requires extra resources, but there are several ways to avoid it. For example, OF can be extracted with additional sub-network~\cite{OFF} or RGB difference~\cite{TSN_rgbdiff_warp} can be used as an alternative motion representation.

%First is so-called two-stream methods which fuse information from multiple modalities, such as RGB and RGB difference, Optical Flow (OF), etc. For example, in \cite{simonyan2014two_stream, Zisserman_Two_Stream_2016} one RGB and 10 OF frames are stacked in order to predict the action at the particular time stamp. In \cite{OFF} authors proposed an advanced fusion mechanism for the extracted Optical Flow and RGB features to achieve state-of-the-art quality. The one evident drawback of this approach is a need to calculate additional modalities, which require extra resources, especially in case of the OF.

Another popular group of methods is related to the use of 3D primitives like 3D Convolution, 3D Batch Normalization, 3D Pooling, and others. They generalize original operations introducing an additional dimension $T$, which indicates the sequence of frames. 
%So that, now the batch is represented by $B \times C \times T \times H \times W$ instead of $B \times C \times H \times W$ and weights of a convolutional kernel are shared across $T$, $H$ and $W$ dimensions, that is why it is called 3D. 
One of the first architectures that leveraged these primitives for the application to AR, is C3D \cite{C3D}. Another famous 3D CNN, which saturated UCF-101 benchmark \cite{UCF101}, is I3D \cite{QuoVadis}. It benefits from pre-training on a large-scale ImageNet \cite{Imagenet} dataset by inflating trained 2D filters into 3D. Although methods based on 3D convolutions allow improving results in terms of accuracy, the computational expenses may achieve dozens of GFLOPs. Another substantial drawback is that at some level of the network only a small number of weights inside the convolutional kernels have a significant impact on the output signal regarding their contribution to the absolute value of activations making utilization of resources ineffective. This problem was mentioned in \cite{CloserLook, xie2018rethinking} where authors proposed decomposition techniques and mixed architectures that combine 3D and 2D operations on different levels of the network.

%Methods of the last category were inspired by NLP and other ML areas where a sequence modeling problem naturally occurs. They utilize recurrent units or attention mechanism \cite{} on top of representation (or embedding) vector which is obtained by a CNN feature extractor from the input image. For example, in \cite{ARUsingVAttn} both a Visual Attention and LSTM cells are used to get a position-wise image embedding and learn temporal dependencies to classify actions. In \cite{HPEAndAR} skeleton features are exploited as an input for attention block to focus on particular parts of the action happening. We argue that these type of methods seem to be less demanding to computational resources because the image embeddings from the previous timestamps can be reused to do a prediction in the future. On the other hand, accuracy of such methods is usually slightly lower than in case of other groups of algorithms.%

%It is not surprising that now hybrid approaches which combine different variations of mentioned methods show the best results and applied in many works \cite{CloserLook, HiddenTwoStream}. In most cases it helps to improve a baseline accuracy and sometimes to simplify a model in order to satisfy a performance criteria.%

%Self attention vs RNN part%

%So, following this tendency we propose an architecture which adopts primitives from CV and NLP \cite{AttnIsAllYouNeed} to get a decent quality on AR task.%

Recurrent neural networks, LSTMs \cite{hochreiter1997long}, and GRUs \cite{chung2014empirical} have been regarded as the default starting point for many sequence modeling problems, such as machine translation or language modeling \cite{goodfellow2016deep}. Many significant results have been achieved in several challenging tasks by means of employing recurrent networks and attention mechanism \cite{shazeer2017outrageously,bahdanau2014neural}. Not surprisingly, several approaches to video classification that model sequences with recurrent connections or gated units have been proposed \cite{yue2015beyond,sharma2016action,donahue2015long}. These models, while showing comparable results on many benchmarks \cite{QuoVadis}, seem to be more suitable for online prediction and thus real-time applications, because feature vector computed for the frame can be reused for predicting classification label for multiple time-windows containing this frame.
% making this method more suitable for real-time applications. 

Several viable alternative approaches to sequence modeling have been proposed recently. These approaches, for example convolutional \cite{bai2018empirical} or fully-attentional (e.g. Transformer~\cite{AttnIsAllYouNeed}) networks, achieve better results on many tasks while addressing significant shortcomings of RNNs such as sequential computing or gradient vanishing. 

We adopt recently proposed Transformer network in our work as a more  elaborate way for sequence modeling. This allowed us to attain high accuracy, retaining the performance, that is sufficient for real-time applications.

\section{Approach}

In this section, we describe a designed approach to AR problem in details as well as discuss some improvements that help to boost the accuracy of our baseline architecture without significantly increasing the complexity. 
% We also give a short comparison of the methods which were reviewed in the previous section on mini-Kinetics dataset, proposed in \cite{xie2018rethinking}. 
% Finally, we discuss a possible area for improvements.

\subsection{Architecture overview}

\begin{figure}[t!]
\includegraphics[width=8cm]{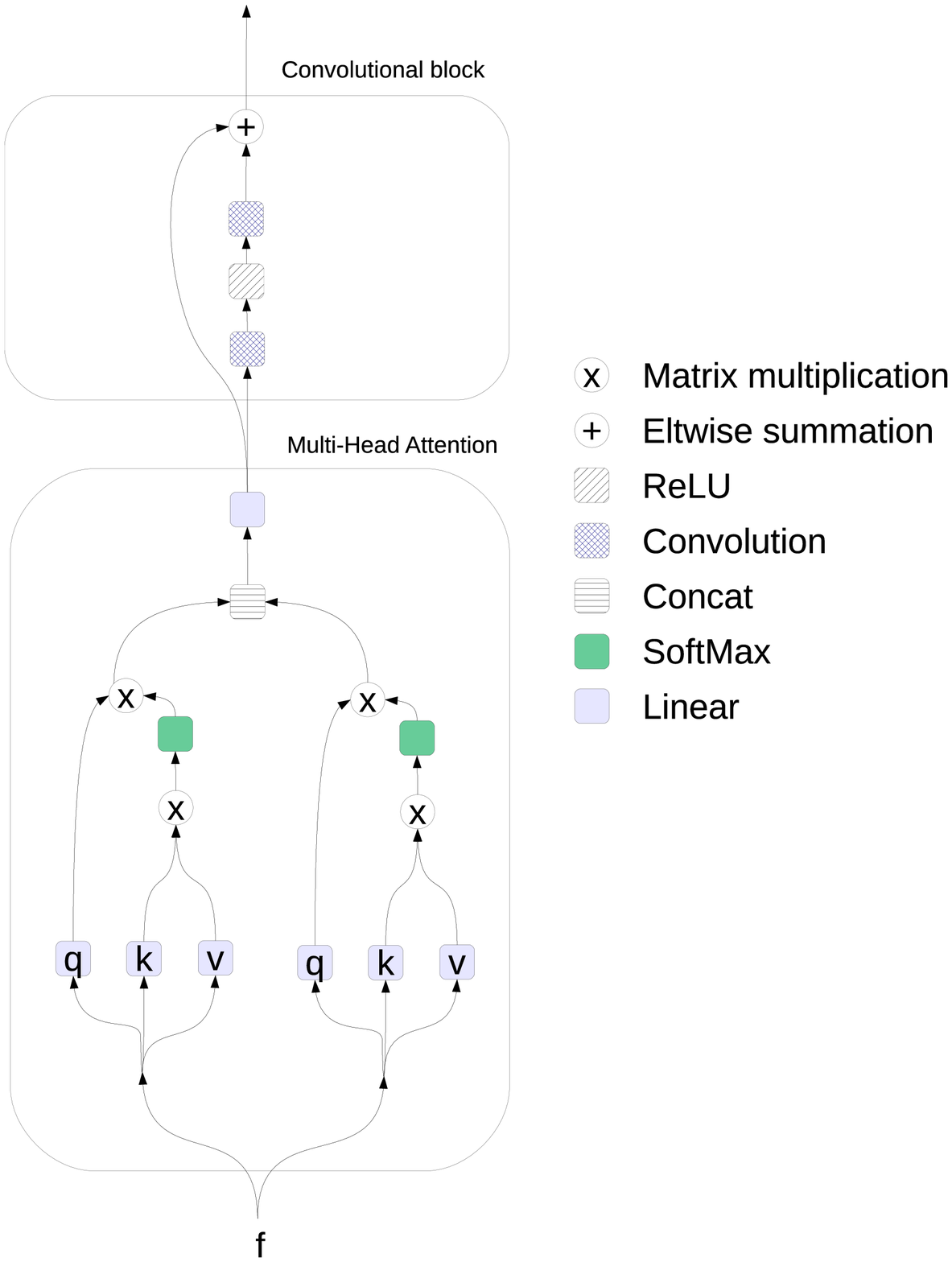}
\caption{The detailed overview of \textbf{decoder block} used in VTN. We use $M = 2$ self-attention heads on the scheme for simplicity. Each head independently transforms input sequence embeddings to its query, key, value triplet using three trainable linear transformations and applies the self-attention operation.  In order to produce output sequence, resulting vectors are concatenated and passed to the block of two convolutions with the kernel of size 1 and residual connection around those convolutions. }
\label{attn_block}
\centering
\end{figure}

% Our method is related to attention mechanism successfully applied to many problems. Specifically, we use approach proposed in \cite{AttnIsAllYouNeed} where so-called Multi-Head Self-Attention blocks were designed to solve a language translation task. The essence of our approach is as follows. 

% We embed each input frame from the sequence to a representation vector of a smaller dimension using a CNN.
% %Herewith, it is not so important what kind of CNN is used for this purpose. 
% For the sake of convenience, we chose ResNet-34 \cite{ResNet} as a baseline architecture in most of our experiments.  

Video Transformer Network (see Fig.~\ref{fig:method_overview}) consists of two parts: the first is the encoder that processes each frame of input sequence independently with 2D CNN in order to get frame embeddings, and the second is the decoder that integrates intra-frame temporal information in a fully-attentional feed-forward fashion, producing the classification label for the given clip. ResNet-34 is used \cite{ResNet} as a baseline architecture for the encoder in most of our experiments. We reuse parameters of all convolutional layers to maximize the benefit of transfer learning from image classification tasks. Global average pooling is then applied to the resulting feature maps to get the frame embeddings of size $d$ (that is equal to 512 in our case), which are then transformed by the decoder, by repeatedly applying multi-head self-attention and convolutional blocks. In multi-head self-attention block, a temporal interrelationship between frames is modeled by informing each frame representation by representation of other frames using the attention mechanism. It consists of several sequential operations. First, vectors of frame representations are mapped to multiple key, value, and query spaces using different learned affine transformations. Each triple of query $Q \in \mathbb{R}^{t \times d_k }$, key $K \in \mathbb{R}^{t \times d_k}$, value $V \in \mathbb{R}^{t \times d_v}$ matrices (where $t$ is the sequence size and $d_k$, $d_v$ are the dimensions of key and value space accordingly) is then transformed to the corresponding head output using the scaled multiplicative attention as following:

\begin{equation}
    \mathrm{head}_i = \mathrm{softmax}(\frac{QK^T}{\sqrt{d_k}})V
\end{equation}

Each head output is then concatenated and passed to the convolutional block that consists of two convolutions with kernel of size 1 (position-wise feedforward) and residual connection. Resulting frame representations are then refined by applying the same procedure multiple times. As we found experimentally, four stacks of such decoder blocks are sufficient for maximizing classification accuracy, and the further increase of the number of blocks did not lead to improvement. In order to produce action confidences for the current clip, a fully-connected layer is applied to all elements of the sequence. Resulting scores are then averaged and normalized with softmax function producing the clip prediction.

\subsection{Multimodal knowledge distillation}

As it was discussed above, the fusion of results of models that receive inputs with different modalities is a common approach to improve the accuracy of Action Recognition algorithm. But in most cases, it leads to a substantial increase in computational complexity due to several reasons. First, it requires to calculate a new modality, which itself may be a hard task, especially in case of the optical flow where commonly used algorithms perform costly iterative energy minimization. Second, since the same architecture is used to do prediction using the second modality, the complexity of the method is doubled. Therefore, both issues make applying of multimodal solutions difficult in real-world applications.

On the other hand, using the RGB difference in place of the optical flow results in almost the same performance \cite{TSN_rgbdiff_warp}, which has been verified by our experiments. At the same time, it requires much lower computational resources that makes using this modality more suitable in conjunction with a still RGB data.

% Knowledge distillation is known to be successful in improving performance of a single model by providing extra supervision signal from ensemble of student networks. However, the question that naturally arises is whether it is possible to transfer knowledge from multiple networks working on different modalities to a single model.%

%In order to avoid using multiple networks for different modalities a knowledge distillation technique \cite{Distillation} seems to be the most suitable. This method allows to transfer knowledge not only from a complex network to a simpler one but also from several networks to one network. However, the question here is whether it is possible to transfer knowledge from CNN that uses different modalities as the input to a CNN that receives only one modality. 

Knowledge distillation \cite{Distillation} is the procedure that designated to help optimization of the student network by providing extra supervision from a larger model or an ensemble of models (teacher). There are successful applications of this technique for reducing the complexity of a larger teacher network \cite{romero2014fitnets} or integrating the performance of an ensemble of models into a single student \cite{bucilua2006model,Distillation}. However, we hypothesize whether it is possible to transfer knowledge from multiple models working on different modalities (two-stream teacher) to a single student.
In order to better understand this, we ran several experiments where knowledge from two ResNet-34 based VTN models working with RGB and RGB difference is distilled to the single RGB model and to the model which receives stacked RGB and RGB difference inputs. We also tried to train a model that operates on stacked input without extra supervision from knowledge distillation. 
Results are shortly summarized in Table~\ref{table:distillation_comparision}. The model working on stacked inputs outperforms the single modality model when trained with knowledge distillation. We suppose that the main reason for that is that motion representation, learned by RGB-difference subnetwork in the two-stream teacher, are not discovered by RGB-only model, yet they significantly contribute to model performance. Note that this technique does not allow matching the performance of the two-stream model. However, it significantly reduces the complexity compared with the original two-stream solution.
% TODO may be provide a comparison with a paper that distilles RGB+OF to RGB?

\setlength{\tabcolsep}{4pt}
\begin{table}
\begin{center}
\caption{Results of knowledge distillation (KD) from two-stream (fusion of two models) ResNet-34-VTN teacher on Mini-Kinetics dataset. The single model that works with stacked modalities improves its accuracy when trained as a student in knowledge distillation setup. However, RGB-only model does not benefit from KD.}
\label{table:distillation_comparision}
\begin{tabular}{lcc}
\hline 
{Model} & {Video@1} & {GMAC\tablefootnote{Billion of multiply-accumulate operations.}}\\
\hline 
Fused RGB + RGB-diff (teacher)  & 78.2 & 7.51\\
RGB  & 75.2 & 3.77 \\
RGB with KD   & 75.2 & 3.77 \\
Stacked RGB + RGB-diff & 75.2 & 3.88 \\
Stacked RGB + RGB-diff with KD & 76.0 & 3.88 \\
\hline
\end{tabular}
\end{center}
\end{table}
\setlength{\tabcolsep}{1.4pt}

\section{Experiments}

In this section we present a study of the proposed method. Kinetics-400 is considered as the primary benchmark. However, the smaller Mini-Kinetics subset that was introduced in \cite{xie2018rethinking} is also used for faster experimentation. We also evaluated our models on UCF-101 and HMDB-51 and evaluated the inference speed on CPU. %and theoretical number of multiply-adds (MACs). 
% We tested our method on several public video classification benchmarks. 

% \subsection{Dataset}
% 

\subsection{Implementation details}

We train and validate our models on 16-frame input sequences that are formed by sampling every second frame from the original video, therefore the total temporal receptive field of our model equals to 32 frames. We tried longer sequences by adding or skipping more frames, but this only resulted in an increased clip accuracy, not the video. In order to calculate video classification accuracy (Video@1), we extracted all non-overlapping 32 frame segments and averaged prediction on these segments.

Frames are scaled in a way, that the shorter side becomes equal to 256. We randomly crop $224 \times 224$ with four different scales during training, as described in \cite{wang2015towards}, and use central $224 \times 224$ crop during the test time. Adam optimizer \cite{kingma2014adam} with the momentum of 0.9 and weight decay of 0.0001 is used. Training is started with the learning rate of $10^{-4}$, which is decayed by a factor of 10 when validation loss reaches a plateau. Models are trained until validation loss stops decreasing, which is usually happened within 50 epochs.

\subsection{Model hyperparameters}

We varied the structure of our decoder block in order to come up with one that maximizes performance on Mini-Kinetics dataset and believed that the same parameter settings would maximize efficiency on other datasets. 

First of all, we evaluated how the number of stacked decoder blocks affects accuracy. We trained models with 1,3,4,5 and 6 blocks, and determined that 4 blocks result in the maximal accuracy and the higher number of blocks does not further boost the metric. We also experimented with sharing parameters between blocks by applying one block recurrently, as suggested in \cite{dehghani2018universal}, but it did not lead to performance improvement. We varied the number of heads in multi-head self-attention, and dimension of query, key $d_k$, and value $d_v$ space, $M = 8$ heads with $d_k = d_v = \frac{d}{M}$ gave the best results. We also tried to add trainable linear transformation after concatenation of heads and to use layer normalization in different locations, but these changes did not affect the accuracy.

\subsection{Comparison with other methods}
\label{sota_compare}

\begin{table*}[!ht]
\centering
\caption{Comparison of different approaches to Action Recognition on Mini-Kinetics dataset with further finetuning on UCF-101 split 1 (Accuracy Video@1). All models are based on the ResNet-34, with the input resolution of 224x224 and 16-frame inputs. Inference time was measured on Intel Core\textsuperscript{TM} i7-8700 CPU @ 2.90GHz and expressed in Frames Per Seconds.}
\label{table:method_comparision}
\begin{tabular*}{\linewidth}{S @{\extracolsep{\fill}}
                        *{5}{S[group-separator = {,},
                               group-minimum-digits = 6
                                ]}}
\toprule
{Model} & {Mini-Kinetics} & {UCF-101} & {MAC} & {FPS} & {Parameters} \\
\hline 
{3D CNN} & {72.9} & {86.4} & {50.2G} & {5}  & {63.5M}\\
{Fused RGB and OF} & {74.3}& {89.8} & {8.5\tablefootnote{Optical flow calculation is not included in the complexity estimation.}} & {32} & {42.8M}\\
{Fused RGB and RGB-diff} & {73.7} & {88.3} & {9.1} & {30} & {42.9M}\\
{Stacked LSTMs} & {72.0}& {86.6} & {3.7} & {55} & {27.6M}\\
{VTN (ours)} & {75.2} & {89.0} & {3.8} & {56} & {29.0M}\\
\bottomrule
\end{tabular*}
\end{table*}

In order to better understand capabilities of the proposed approach, we compare it with methods described in Section~\ref{RelatedWorks}. For a fair comparison, we take ResNet-34 architecture and extend it to the case of 3D networks and two-stream methods in the way described below.

The first model we compare with is ResNet-34 3D which is described in \cite{3DResNets}. It repeats a common ResNet architecture, but instead of 2D Convolutions and Pooling layers, it utilizes their 3D analogs. A global Average Pooling operation over three dimensions is applied at the end of the network in order to get a representation vector, which is fed to a fully-connected layer producing the CNN output. Vanilla ResNet-34 pre-trained on ImageNet is used to initialize its 3D analog where convolutional kernels are repeated over temporal dimension $T$, as proposed in \cite{QuoVadis}.

The next approach that we consider is a two-stream model that is represented by a fusion of two ResNet-34 CNNs trained on RGB and OF inputs. The OF model is almost the original ResNet-34, but its first convolutional layer receives 32-channels input, formed by $X$ and $Y$ components of pre-calculated optical flow for 16 sequential frames. To initialize this layer we average the first convolutional kernel of the RGB model pre-trained on ImageNet over the channel dimension and repeat it 32 times. 

We also tried a two-stream model where two fused CNNs were trained on RGB and RGB difference inputs since the calculation of the latter is much cheaper than the optical flow. In this case, the motion model receives 48-channels input of RGB differences from 16 consecutive frames.

The last model examined in our comparison is the ResNet-34 followed by three stacked LSTM cells operating on independent frame embeddings. As before, we use the ImageNet pre-trained model for initialization, but learn LSTM parameters from scratch. We found this model quite simple but representative at the same time. We also tried to apply a visual attention mechanism, as suggested in \cite{sharma2016action}, but it did not improve the performance. 

The comparison of the described models and our proposed method is shown in Table~\ref{table:method_comparision}. For the sake of convenience, we also provide a theoretical complexity and inference time for all models. The input resolution is set to 224x224, and the sequence size is 16 frames for all models. The models were trained with Adam optimizer until validation loss reaches the plateau. The obtained results show that our VTN model outperforms others on Mini-Kinetics dataset and works on par with the two-stream method. We find this fact surprising because we believe that 3D Convolutional model should perform better because it consists of operations that can learn temporal dependencies at every layer and has a higher capacity regarding the number of parameters. 
%We assume that a reason for that may be in the initialization process because other non-3D models have the strong baselines for their backbones which are based on ImageNet pre-trained models. In the case of 3D models such initialization is not quite correct since all kernels along time dimension initialized by the same weights.
%We hypothesize that here may be two reasons. The first is that other non-3D models have the strong baselines for their backbones which are based on ImageNet pre-trained models. In the case of 3D models, such initialization is not quite correct since all kernels along time dimension have the same weights. The second reason is related to the learning process. Since the 3D model has much more parameters, the convergence to the best optimum can be difficult.

Another interesting result is that the two-stream RGB-difference model shows the performance that is close to the OF-based model while saving a large number of calculations.
These findings correspond to the results of \cite{3DResNets, RepresentationFlow}. Nevertheless, our VTN approach is attractive in terms of speed/accuracy trade-off.

\subsection{Comparison with state-of-the-art}

\setlength{\tabcolsep}{4pt}
\begin{table}
\begin{center}
\caption{Comparison with the state-of-the-art on \textbf{Kinetics-400} dataset.}
\label{table:ComparisionWithOthers}
\begin{tabular}{lcc}
\hline\noalign{\smallskip}
Method & Video@1 \\
\noalign{\smallskip}
\hline
\noalign{\smallskip}
BNInception+TSN-RGB \cite{TSN_rgbdiff_warp} & 69.1\tablefootnote{Author's implementation (https://github.com/yjxiong/tsn-pytorch) uses 10-crop TTA during testing.}\\
I3D-RGB \cite{QuoVadis} & 72.1\\
I3D-TwoStream \cite{QuoVadis} & 75.7\\
S3D-G \cite{xie2018rethinking} & 74.7\\
R(2+1)D-TwoStream \cite{CloserLook} & 75.4\\
R(2+1)D-RGB \cite{CloserLook} & 74.3\\
NL-I3D-ResNet-101-RGB \cite{wang2017non} & 77.7\\
\midrule
MobileNetV2-VTN-RGB & 62.5\\
ResNet-34-VTN-RGB & 68.3\\
ResNet-34-VTN-RGB+RGBDiff & 71.0\\
SE-ResNeXt-101-VTN-RGB & 69.5\\
SE-ResNeXt-101-VTN-RGB+RGDiff & 73.5\\
\hline
\end{tabular}
\end{center}
\end{table}
\setlength{\tabcolsep}{1.4pt}

To compare with other state-of-the-art models we assessed our approach on Kinetics-400 dataset. In addition to the baseline ResNet-34-VTN, we used a larger model employing SE-ResNeXt-101 (32x4d) architecture for the encoder, which is, however, still very cheap in terms of a number of multiply-accumulates in comparison with 3D CNNs. Another interesting question is the potential of the proposed method in optimizing a model for mobile devices and what associated drop in accuracy it would incur. To tackle this question we tested our approach with the lightweight MobileNetV2 \cite{sandler2018mobilenetv2} encoder. 

Since fusion of prediction from streams with different modalities (e.g. RGB and optical flow or RGB and RGB difference) allowed improving results in many published works, we experimented with enhancing the results of our RGB model by combining it with the analogous RGB difference model. We subtracted normalized adjacent frames and trained the ResNet34-VTN model on this data. This allowed us to improve the results of the ResNet34-VTN model by a margin of 2.4\%. 

The results for the Kinetics-400 validation set are presented in Table~\ref{table:ComparisionWithOthers}. The breakthrough I3D model \cite{QuoVadis} outperforms ResNet-34 VTN and SE-ResNeXt-101 (32x4d) VTN only by a small margin of 3.5\% and 2.1\% accordingly, thus our method still shows competitive results while being computationally significantly cheaper for online prediction scenarios. 

We also provide results on the popular UCF-101 and HMDB-51 datasets. We fine-tuned models trained on Kinetics-400 for 20 epochs with smaller learning rate of $10^{-5}$. Mean video accuracies over three validation splits are presented in Table~\ref{table:sota_ucf}.

% We finetuned on UCF...

% We also briefly tried to combine our method with the Two-Stream approach by training analogous model, but working on optical flow (RGB difference) modality. ...

Computational complexity versus accuracy on Kinetics-400 for some state-of-the-art methods and various variants of VTN is shown in Fig.~\ref{tradeoff}. 
Since we primarily focus on the online prediction scenario (i.e. when the classification label is required for every subsequent frame) we consider the number of operations needed to execute the encoder on one frame as well as operations for the whole decoder. On the other hand, 3D convolutional models extract features from adjacent frames and require to execute the entire network for each new frame. Thus our method is more attractive in terms of accuracy/complexity for real-time applications.
% \subsection{Results}
% what to add in this section?

\setlength{\tabcolsep}{4pt}
\begin{table}
\begin{center}
\caption{Comparison with other methods on \textbf{UCF-101} and \textbf{HMDB-51} (average metric over all splits). Methods of the first set of rows do not use Kinetics pre-training.}
\label{table:sota_ucf}
\begin{tabular}{lcc}
\hline\noalign{\smallskip}
Method & UCF-101 & HMDB-51 \\
\noalign{\smallskip}
\hline
\noalign{\smallskip}

IDT \cite{wang2013action} & 86.4 & 61.7 \\
C3D \cite{C3D} & 85.2 & - \\
Two-Stream \cite{simonyan2014two_stream} & 88.0 & 59.4 \\
Two-Stream Fusion + IDT \cite{feichtenhofer2016convolutional} & 93.5 & 69.2 \\
BNInception+TSN-RGB \cite{TSN_rgbdiff_warp} & 91.1 & - \\
P3D \cite{TSN_rgbdiff_warp} & 88.6 & - \\
ST-ResNet + IDT \cite{feichtenhofer2016spatiotemporal} & 94.6 & 70.3 \\
\midrule
I3D-RGB \cite{QuoVadis} & 95.6  & 74.8 \\
I3D-TwoStream \cite{QuoVadis} & 98.0 & 80.7 \\
S3D-G \cite{xie2018rethinking} & 96.8 & 75.9 \\
R(2+1)D-TwoStream \cite{CloserLook} & 97.3 & 78.7\\
\midrule
ResNet-34-VTN-RGB  & 90.8 & 63.5 \\
SE-ResNeXt-101-VTN-RGB & 92.2 & 67.2 \\
ResNet-34-VTN-RGB+RGBDiff  & 95.0 & 71.3 \\
\makecell[l]{SE-ResNeXt-101-VTN-RGB+\\RGBDiff}  & 95.0 & 71.6 \\
\hline
\end{tabular}
\end{center}
\end{table}
\setlength{\tabcolsep}{1.4pt}

\subsection{Inference speed}
Since theoretically faster models do not necessarily correspond to higher inference speed \cite{ma2018shufflenet,liu2017learning,sandler2018inverted}, we also evaluate the actual inference time to prove the feasibility of the proposed method for real-time applications. Currently, there are several frameworks available, such as Nvidia Tensor RT \cite{TensorRT} or Intel\textsuperscript{\textregistered{}} OpenVINO\textsuperscript{TM} Toolkit~\cite{OpenVINO}, which can highly optimize DL model for particular hardware. Since we primarily focus on models suitable for edge computing, we chose OpenVINO and its DL Deployment Toolkit as the inference engine for our solution. OpenVINO can import models from many DL frameworks as well as ONNX \cite{ONNX} representation which we use to convert models from PyTorch framework which is used in all our experiments.

\setlength{\tabcolsep}{4pt}
\begin{table}
\begin{center}
\caption{\textbf{Inference time} of various Video Transformer Networks with OpenVINO on Intel Core\textsuperscript{TM} i7-8700 CPU @ 2.90GHz.}
\label{table:TargetPerformanceResults}
\begin{tabular}{lcc}
\hline\noalign{\smallskip}
Model & FPS & GMAC\\
\noalign{\smallskip}
\hline
\noalign{\smallskip}
ResNet-34-VTN-RGB & 56 & 3.77\\
\makecell[l]{Stacked RGB+RGBDiff \\ ResNet-34 VTN} & 51 &4.2\\
ResNet-50-VTN-RGB & 49 & 4.25 \\
MobileNetV2-VTN-RGB & 177 & 0.4 \\
\hline
\end{tabular}
\end{center}
\end{table}
\setlength{\tabcolsep}{1.4pt}

Table~\ref{table:TargetPerformanceResults} shows the inference time on CPU of several models that employ the proposed approach. Faster than real-time speed is achieved for all models, making this method promising for edge computing.

\section{Conclusions}
In this work, we have proposed a new Video Transformer Network architecture for real-time Action Recognition. We have shown that adopting methods from Natural Language Processing along with using an appropriate CNN for Image Classification helps to achieve accuracy on-par with state-of-the-art methods. Moreover, it has been demonstrated that the proposed approach favorably compares with other approaches, such as 3D Convolution-based models or two-stream methods. Specifically, it allows utilizing computational resources more effectively by embedding each input frame to lower-dimensional high-level feature vector and then making a conclusion about the action operating only on embedding vectors by means of self-attention. This method allows achieving real-time inference on a general-purpose CPU, providing capabilities for using AR algorithms at the edge. Our research also demonstrates that the self-attention mechanism is quite universal and can be applied to many tasks, such as Natural Language Processing, Speech Recognition or Computer Vision.

%This also may be considered as an additional proof that the high-level information about the observing scene is more important than a low level one to make an accurate prediction about the whole video sequence.

{\small
\bibliographystyle{ieee}
\bibliography{arxiv_version}

\begin{thebibliography}{10}\itemsep=-1pt

\bibitem{TensorRT}
{NVIDIA TensorRT Programmable Inference Accelerator}.
\newblock \url{https://developer.nvidia.com/tensorrt}.

\bibitem{ONNX}
{ONNX}.
\newblock \url{https://onnx.ai/}.

\bibitem{OpenVINO}
{OpenVINO Toolkit}.
\newblock \url{https://software.intel.com/en-us/openvino-toolkit}.

\bibitem{bahdanau2014neural}
D.~Bahdanau, K.~Cho, and Y.~Bengio.
\newblock Neural machine translation by jointly learning to align and
  translate.
\newblock {\em arXiv preprint arXiv:1409.0473}, 2014.

\bibitem{bai2018empirical}
S.~Bai, J.~Z. Kolter, and V.~Koltun.
\newblock An empirical evaluation of generic convolutional and recurrent
  networks for sequence modeling.
\newblock {\em arXiv preprint arXiv:1803.01271}, 2018.

\bibitem{bucilua2006model}
C.~Bucilua, R.~Caruana, and A.~Niculescu-Mizil.
\newblock Model compression.
\newblock In {\em Proceedings of the 12th ACM SIGKDD international conference
  on Knowledge discovery and data mining}, pages 535--541. ACM, 2006.

\bibitem{OpenPose}
Z.~Cao, T.~Simon, S.-E. Wei, and Y.~Sheikh.
\newblock Realtime multi-person 2d pose estimation using part affinity fields.
\newblock {\em arXiv preprint arXiv:1611.08050}, 2016.

\bibitem{massivelyParallel}
J.~Carreira, V.~Patraucean, L.~Mazare, A.~Zisserman, and S.~Osindero.
\newblock Massively parallel video networks.
\newblock In {\em The European Conference on Computer Vision (ECCV)}, September
  2018.

\bibitem{QuoVadis}
J.~Carreira and A.~Zisserman.
\newblock Quo vadis, action recognition? a new model and the kinetics dataset.
\newblock In {\em Computer Vision and Pattern Recognition (CVPR), 2017 IEEE
  Conference on}, pages 4724--4733. IEEE, 2017.

\bibitem{Xception}
F.~Chollet.
\newblock Xception: Deep learning with depthwise separable convolutions.
\newblock In {\em CVPR}, pages 1251--1258, 2017.

\bibitem{chung2014empirical}
J.~Chung, C.~Gulcehre, K.~Cho, and Y.~Bengio.
\newblock Empirical evaluation of gated recurrent neural networks on sequence
  modeling.
\newblock {\em arXiv preprint arXiv:1412.3555}, 2014.

\bibitem{Cityscapes}
M.~Cordts, M.~Omran, S.~Ramos, T.~Rehfeld, M.~Enzweiler, R.~Benenson,
  U.~Franke, S.~Roth, and B.~Schiele.
\newblock The cityscapes dataset for semantic urban scene understanding.
\newblock In {\em Proceedings of the IEEE conference on computer vision and
  pattern recognition}, pages 3213--3223, 2016.

\bibitem{dehghani2018universal}
M.~Dehghani, S.~Gouws, O.~Vinyals, J.~Uszkoreit, and {\L}.~Kaiser.
\newblock Universal transformers.
\newblock {\em arXiv preprint arXiv:1807.03819}, 2018.

\bibitem{Imagenet}
J.~Deng, W.~Dong, R.~Socher, L.-J. Li, K.~Li, and L.~Fei-Fei.
\newblock Imagenet: A large-scale hierarchical image database.
\newblock In {\em CVPR}, 2009.

\bibitem{donahue2015long}
J.~Donahue, L.~Anne~Hendricks, S.~Guadarrama, M.~Rohrbach, S.~Venugopalan,
  K.~Saenko, and T.~Darrell.
\newblock Long-term recurrent convolutional networks for visual recognition and
  description.
\newblock In {\em Proceedings of the IEEE conference on computer vision and
  pattern recognition}, pages 2625--2634, 2015.

\bibitem{VOC}
M.~Everingham, L.~V. Gool, C.~K. Williams, J.~Winn, and A.~Zisserman.
\newblock The pascal visual object classes (voc) challenge.
\newblock {\em International Journal of Computer Vision}, 88(2):303--338, 2010.

\bibitem{feichtenhofer2016spatiotemporal}
C.~Feichtenhofer, A.~Pinz, and R.~Wildes.
\newblock Spatiotemporal residual networks for video action recognition.
\newblock In {\em Advances in neural information processing systems}, pages
  3468--3476, 2016.

\bibitem{Zisserman_Two_Stream_2016}
C.~Feichtenhofer, A.~Pinz, and A.~Zisserman.
\newblock Convolutional two-stream network fusion for video action recognition.
\newblock In {\em Proceedings of the IEEE Conference on Computer Vision and
  Pattern Recognition}, pages 1933--1941, 2016.

\bibitem{feichtenhofer2016convolutional}
C.~Feichtenhofer, A.~Pinz, and A.~Zisserman.
\newblock Convolutional two-stream network fusion for video action recognition.
\newblock In {\em Proceedings of the IEEE Conference on Computer Vision and
  Pattern Recognition}, pages 1933--1941, 2016.

\bibitem{goodfellow2016deep}
I.~Goodfellow, Y.~Bengio, A.~Courville, and Y.~Bengio.
\newblock {\em Deep learning}, volume~1.
\newblock MIT press Cambridge, 2016.

\bibitem{3DResNets}
K.~Hara, H.~Kataoka, and Y.~Satoh.
\newblock Can spatiotemporal 3d cnns retrace the history of 2d cnns and
  imagenet.
\newblock In {\em Proceedings of the IEEE Conference on Computer Vision and
  Pattern Recognition, Salt Lake City, UT, USA}, pages 18--22, 2018.

\bibitem{MaskR-CNN}
K.~He, G.~Gkioxari, P.~Doll{\'a}r, and R.~Girshick.
\newblock Mask r-cnn.
\newblock In {\em Computer Vision (ICCV), 2017 IEEE International Conference
  on}, pages 2980--2988. IEEE, 2017.

\bibitem{ResNet}
K.~He, X.~Zhang, S.~Ren, and J.~Sun.
\newblock Deep residual learning for image recognition.
\newblock In {\em Proceedings of the IEEE conference on computer vision and
  pattern recognition}, pages 770--778, 2016.

\bibitem{Distillation}
G.~Hinton, O.~Vinyals, and J.~Dean.
\newblock Distilling the knowledge in a neural network, 2015.

\bibitem{hochreiter1997long}
S.~Hochreiter and J.~Schmidhuber.
\newblock Long short-term memory.
\newblock {\em Neural computation}, 9(8):1735--1780, 1997.

\bibitem{MobileNet}
A.~G. Howard, M.~Zhu, B.~Chen, D.~Kalenichenko, W.~Wang, T.~Weyand,
  M.~Andreetto, and H.~Adam.
\newblock Mobilenets: Efficient convolutional neural networks for mobile vision
  applications. arxiv:1704.04861, 2017.

\bibitem{SqueezeNet}
F.~N. Iandola, S.~Han, M.~W. Moskewicz, K.~Ashraf, W.~J. Dally, and K.~Keutzer.
\newblock Squeezenet: Alexnet-level accuracy with 50x fewer parameters and< 0.5
  mb model size.
\newblock {\em arXiv preprint arXiv:1602.07360}, 2016.

\bibitem{Kinetics}
W.~Kay, J.~Carreira, K.~Simonyan, B.~Zhang, C.~Hillier, S.~Vijayanarasimhan,
  F.~Viola, T.~Green, T.~Back, P.~Natsev, et~al.
\newblock The kinetics human action video dataset.
\newblock {\em arXiv preprint arXiv:1705.06950}, 2017.

\bibitem{kingma2014adam}
D.~P. Kingma and J.~Ba.
\newblock Adam: A method for stochastic optimization.
\newblock {\em arXiv preprint arXiv:1412.6980}, 2014.

\bibitem{AlexNet}
A.~Krizhevsky, I.~Sutskever, and G.~E. Hinton.
\newblock Imagenet classification with deep convolutional neural networks.
\newblock In F.~Pereira, C.~J.~C. Burges, L.~Bottou, and K.~Q. Weinberger,
  editors, {\em Advances in Neural Information Processing Systems 25}, pages
  1097--1105. Curran Associates, Inc., 2012.

\bibitem{HMDB}
H.~Kuehne, H.~Jhuang, E.~Garrote, T.~Poggio, and T.~Serre.
\newblock Hmdb: a large video database for human motion recognition.
\newblock In {\em Computer Vision (ICCV), 2011 IEEE International Conference
  on}, pages 2556--2563. IEEE, 2011.

\bibitem{MSCOCO}
T.-Y. Lin, M.~Maire, S.~Belongie, J.~Hays, P.~Perona, D.~Ramanan,
  P.~Doll\'{a}r, and C.~L. Zitnick.
\newblock Microsoft coco: Common objects in context.
\newblock In {\em ECCV}, pages 740--755, 2014.

\bibitem{liu2017learning}
Z.~Liu, J.~Li, Z.~Shen, G.~Huang, S.~Yan, and C.~Zhang.
\newblock Learning efficient convolutional networks through network slimming.
\newblock In {\em Computer Vision (ICCV), 2017 IEEE International Conference
  on}, pages 2755--2763. IEEE, 2017.

\bibitem{ma2018shufflenet}
N.~Ma, X.~Zhang, H.-T. Zheng, and J.~Sun.
\newblock Shufflenet v2: Practical guidelines for efficient cnn architecture
  design.
\newblock {\em arXiv preprint arXiv:1807.11164}, 2018.

\bibitem{RepresentationFlow}
A.~Piergiovanni and M.~S. Ryoo.
\newblock Representation flow for action recognition, 2018.

\bibitem{romero2014fitnets}
A.~Romero, N.~Ballas, S.~E. Kahou, A.~Chassang, C.~Gatta, and Y.~Bengio.
\newblock Fitnets: Hints for thin deep nets.
\newblock {\em arXiv preprint arXiv:1412.6550}, 2014.

\bibitem{sandler2018inverted}
M.~Sandler, A.~Howard, M.~Zhu, A.~Zhmoginov, and L.-C. Chen.
\newblock Inverted residuals and linear bottlenecks: Mobile networks for
  classification, detection and segmentation.
\newblock {\em arXiv preprint arXiv:1801.04381}, 2018.

\bibitem{sandler2018mobilenetv2}
M.~Sandler, A.~Howard, M.~Zhu, A.~Zhmoginov, and L.-C. Chen.
\newblock Mobilenetv2: Inverted residuals and linear bottlenecks.
\newblock In {\em Proceedings of the IEEE Conference on Computer Vision and
  Pattern Recognition}, pages 4510--4520, 2018.

\bibitem{sharma2016action}
S.~Sharma, R.~Kiros, and R.~Salakhutdinov.
\newblock Action recognition using visual attention.
\newblock 2016.

\bibitem{shazeer2017outrageously}
N.~Shazeer, A.~Mirhoseini, K.~Maziarz, A.~Davis, Q.~Le, G.~Hinton, and J.~Dean.
\newblock Outrageously large neural networks: The sparsely-gated
  mixture-of-experts layer.
\newblock {\em arXiv preprint arXiv:1701.06538}, 2017.

\bibitem{simonyan2014two_stream}
K.~Simonyan and A.~Zisserman.
\newblock Two-stream convolutional networks for action recognition in videos.
\newblock In {\em Advances in neural information processing systems}, pages
  568--576, 2014.

\bibitem{VGG}
K.~Simonyan and A.~Zisserman.
\newblock Very deep convolutional networks for large-scale image recognition,
  2014.

\bibitem{UCF101}
K.~Soomro, A.~R. Zamir, and M.~Shah.
\newblock Ucf101: A dataset of 101 human actions classes from videos in the
  wild.
\newblock {\em arXiv preprint arXiv:1212.0402}, 2012.

\bibitem{OFF}
S.~Sun, Z.~Kuang, L.~Sheng, W.~Ouyang, and W.~Zhang.
\newblock Optical flow guided feature: A fast and robust motion representation
  for video action recognition.
\newblock In {\em The IEEE Conference on Computer Vision and Pattern
  Recognition (CVPR)(June 2018)}, 2018.

\bibitem{InceptionV1}
C.~Szegedy, W.~Liu, Y.~Jia, P.~Sermanet, S.~Reed, D.~Anguelov, D.~Erhan,
  V.~Vanhoucke, and A.~Rabinovich.
\newblock Going deeper with convolutions.
\newblock In {\em Proceedings of the IEEE conference on computer vision and
  pattern recognition}, pages 1--9, 2015.

\bibitem{C3D}
D.~Tran, L.~Bourdev, R.~Fergus, L.~Torresani, and M.~Paluri.
\newblock Learning spatiotemporal features with 3d convolutional networks.
\newblock In {\em Proceedings of the IEEE international conference on computer
  vision}, pages 4489--4497, 2015.

\bibitem{CloserLook}
D.~Tran, H.~Wang, L.~Torresani, J.~Ray, Y.~LeCun, and M.~Paluri.
\newblock A closer look at spatiotemporal convolutions for action recognition.
\newblock In {\em Proceedings of the IEEE Conference on Computer Vision and
  Pattern Recognition}, pages 6450--6459, 2018.

\bibitem{AttnIsAllYouNeed}
A.~Vaswani, N.~Shazeer, N.~Parmar, J.~Uszkoreit, L.~Jones, A.~N. Gomez,
  {\L}.~Kaiser, and I.~Polosukhin.
\newblock Attention is all you need.
\newblock In {\em Advances in Neural Information Processing Systems}, pages
  5998--6008, 2017.

\bibitem{wang2013action}
H.~Wang and C.~Schmid.
\newblock Action recognition with improved trajectories.
\newblock In {\em Proceedings of the IEEE international conference on computer
  vision}, pages 3551--3558, 2013.

\bibitem{wang2015towards}
L.~Wang, Y.~Xiong, Z.~Wang, and Y.~Qiao.
\newblock Towards good practices for very deep two-stream convnets.
\newblock {\em arXiv preprint arXiv:1507.02159}, 2015.

\bibitem{TSN_rgbdiff_warp}
L.~Wang, Y.~Xiong, Z.~Wang, Y.~Qiao, D.~Lin, X.~Tang, and L.~Van~Gool.
\newblock Temporal segment networks: Towards good practices for deep action
  recognition.
\newblock In {\em European Conference on Computer Vision}, pages 20--36.
  Springer, 2016.

\bibitem{wang2017non}
X.~Wang, R.~Girshick, A.~Gupta, and K.~He.
\newblock Non-local neural networks.
\newblock {\em arXiv preprint arXiv:1711.07971}, 10, 2017.

\bibitem{xie2018rethinking}
S.~Xie, C.~Sun, J.~Huang, Z.~Tu, and K.~Murphy.
\newblock Rethinking spatiotemporal feature learning: Speed-accuracy trade-offs
  in video classification.
\newblock In {\em Proceedings of the European Conference on Computer Vision
  (ECCV)}, pages 305--321, 2018.

\bibitem{yue2015beyond}
J.~Yue-Hei~Ng, M.~Hausknecht, S.~Vijayanarasimhan, O.~Vinyals, R.~Monga, and
  G.~Toderici.
\newblock Beyond short snippets: Deep networks for video classification.
\newblock In {\em Proceedings of the IEEE conference on computer vision and
  pattern recognition}, pages 4694--4702, 2015.

\bibitem{TVL1}
C.~Zach, T.~Pock, and H.~Bischof.
\newblock A duality based approach for realtime tv-l 1 optical flow.
\newblock In {\em Joint Pattern Recognition Symposium}, pages 214--223.
  Springer, 2007.

\bibitem{ShuffleNet}
X.~Zhang, X.~Zhou, M.~Lin, and J.~Sun.
\newblock Shufflenet: An extremely efficient convolutional neural network for
  mobile devices, 2017.

\bibitem{PSPNet}
H.~Zhao, J.~Shi, X.~Qi, X.~Wang, and J.~Jia.
\newblock Pyramid scene parsing network.
\newblock In {\em IEEE Conf. on Computer Vision and Pattern Recognition
  (CVPR)}, pages 2881--2890, 2017.

\end{thebibliography}
}

\end{document}